# Retail Demand Forecasting: A Comparative Study for Multivariate Time Series

Md Sabbirul Haque
Institute of Electrical and Electronics Engineers
sabbir465@ieee.org

Md Shahedul Amin
University of Tennessee Chattanooga
Department Finance & Economics
shahedul.amin94@gmail.com

Jonayet Miah
Department of Computer Science
University of South Dakota
jonayet.miah@coyotes.usd.edu

## Abstract

Accurate demand forecasting in the retail industry is a critical determinant of financial performance and supply chain efficiency. As global markets become increasingly interconnected, businesses are turning towards advanced prediction models to gain a competitive edge. However, existing literature mostly focuses on historical sales data and ignores the vital influence of macroeconomic conditions on consumer spending behavior. In this study, we bridge this gap by enriching time series data of customer demand with macroeconomic variables, such as the Consumer Price Index (CPI), Index of Consumer Sentiment (ICS), and unemployment rates. Leveraging this comprehensive dataset, we develop and compare various regression and machine learning models to predict retail demand accurately.

# 1. Introduction

In today's rapidly evolving and fiercely competitive retail landscape, the precision of demand forecasting has become paramount. The ability to accurately predict demand not only facilitates optimized inventory management and cost reduction but also serves as a catalyst for elevating customer satisfaction. However, relying solely on traditional forecasting techniques that hinge on historical sales data falls short of capturing the intricate dynamics that are influenced by a multitude of external economic factors. Current studies in the literature delve into the exploration of advanced regression and machine learning algorithms, seeking methodologies to enhance predictive accuracy. While these advanced algorithms harness the potency of historical sales data, they mostly overlook the profound impact of macroeconomic conditions on the ever-shifting behavior of household shopping.

Within the scholarly discourse, documented empirical evidence demonstrates that macroeconomic variables display substantive influence over consumer expenditure trends. In particular, the Consumer Price Index (CPI), Index of Consumer Sentiment (ICS), and unemployment rates manifest as pivotal factors in explaining retail demand. However, within the realm of demand forecasting efforts, the evidence of capturing this influence of external economic factors in shaping households' demand patterns is conspicuously scanty. This realization serves as a driving force for our research, as we strengthen the weakly established finding in the current literature of superior performance of retail forecasting for integrating these macroeconomic factors with time series data, and further confirm this weakly established finding by applying various regression and machine learning models in solving retail forecasting problems. Going beyond the realm of refining retail demand forecasting with amplified precision, our academic endeavor extends to the inherent limitations of conventional approaches. The integration of macroeconomic indicators as integral constituents of our forecasting framework enables us to capture intricate relations between economic conditions and consumer behaviors.

In short, within the context of a constantly evolving retail arena characterized by fierce competition, the spotlight upon the precision of demand forecasting intensifies. Bearing in mind the constraints intrinsic to methodologies tethered solely to historical data, our scholarly focus lies in the augmentation of time series data through the seamless integration of macroeconomic variables, thereby confirming the pivotal role these factors play in shaping the contours of future retail demand. Our research aims to improve this by using economic indicators alongside the usual data to predict demand better and understand the connection between economics and shopping habits. It's about going beyond just improving predictions – it's about uncovering a deeper understanding of how economic conditions affect what people buy.

# 2. Literature Review

Accurate demand forecasting is imperative for effective supply chain management, guiding critical decisions from production planning to inventory optimization. This review synthesizes key findings on demand forecasting approaches, documenting the shift from statistical methods to advanced machine learning techniques. Additionally, the integration of macroeconomic and consumer indicators is highlighted as an emerging approach to enrich predictive models and enhance retail sales forecasting accuracy. A substantial body of research has explored modeling techniques, with initial dominance of statistical methods, such as SARIMA but a recent shift towards machine learning approaches. SARIMA has been widely adopted given its maturity and interpretability as a time series technique capturing both seasonal and trend components (Abellana et al., 2020). Chuang (1991) developed a SARIMA model for computer part demand forecasting that improved on simple smoothing methods. More recently,

Velos et al. (2020) successfully applied SARIMA for sales forecasting in the alcoholic beverage industry, finding it outperformed baseline models. However, as data complexity grows, limitations of linear modeling have become apparent.

In response to the constraints of statistical techniques, machine learning has emerged as a flexible and powerful approach for modeling complex nonlinear relationships present in demand data. Makridakis et al. (2018) benchmarked statistical methods against machine learning, finding Random Forests algorithms reduced forecast errors. Bousqaoui et al. (2021) showed Support Vector Regression significantly improved upon SARIMA and Exponential Smoothing Forecasts. Additionally, combinations of machine learning techniques have proven effective. Mitra and Sahu (2022) conduct a comparative study of five regression techniques in machine learning, including Random Forests, Extreme Gradient Boosting, Gradient Boosting, Adaptive Boosting, and Artificial Neural Networks algorithms. Their findings reveal that a hybrid model combining Random Forests, Extreme Gradient Boosting, and linear regression exhibits superior forecasting accuracy compared to the other models, indicating the potential of ensemble methods in demand forecasting. Lorente-Leyva et al. (2020) undertake a comparison of machine learning techniques, such as Artificial Neural Networks, with classical demand forecasting methods in the Ecuadorian textile industry. Their study demonstrates that machine learning techniques outperform classical methods in terms of accuracy and overall performance, highlighting the advantages of leveraging advanced algorithms in demand forecasting applications.

Mbonyinshuti and Kim (2021) apply machine learning algorithms, such as linear regression, Artificial Neural Networks, and Random Forest, to predict the demand for essential medicines in Rwanda using consumption data. Their research showcases the effectiveness of machine learning approaches in improving demand forecasting accuracy, potentially contributing to better healthcare resource planning and management. Punia et al. (2020) observe that traditional time-series methods, such as exponential smoothing and ARIMA, were commonly used for demand forecasting in offline retail. However, their study suggests that the new proposition they introduce outperforms these traditional methods, highlighting the potential benefits of adopting advanced machine learning approaches in demand forecasting.

Grzywińska-Rąpca and Ptak-Chmielewska (2023) investigate the determinants of the Consumer Confidence Index (CCI) and its influence on consumer spending behavior. The study emphasizes the role of respondent expectations in shaping the CCI and demonstrates its significance in forecasting consumption and retail sales. By analyzing consumer confidence as a leading indicator, the authors shed light on its potential contribution to retail sales forecasting accuracy. Ye and Le (2023) delve into the factors influencing automobile sales, with particular attention to macroeconomic determinants. Their study presents an automobile sales forecasting model that incorporates variables such as residents' disposable income and other relevant macroeconomic indicators. The integration of such factors allows for a more comprehensive understanding of the dynamics driving automobile sales and improves forecasting precision.

It has been documented in the literature that macroeconomic situation can affect individual behaviour [Haque 2020]. Bakas and Triantafyllou (2019) explore the predictive power of macroeconomic uncertainty on the volatility of agricultural, energy, and metals commodity markets. By examining the relationship between macroeconomic indicators and commodity market volatility, the authors highlight the crucial role that macroeconomic variables play in forecasting market fluctuations. Their findings suggest that incorporating macroeconomic uncertainty enhances the accuracy of volatility forecasts in commodity markets. Liu et al. (2022) propose a regional economic forecasting method based on recurrent neural networks. Given the nonlinear and uncertain nature of macroeconomic systems, the authors advocate for the adoption of advanced machine learning techniques in regional economic forecasting. By leveraging recurrent neural networks, they demonstrate the potential to capture complex relationships between macroeconomic and consumer indicators, thus improving the precision of regional sales forecasts.

In his paper, Haque (2023) introduces the integration of external macroeconomic variables, such as the Consumer Price Index (CPI), Consumer Sentiment Index (ICS), and unemployment rates—in conjunction with time series data related to retail product sales. The purpose of this integration is to enrich the training dataset fed into a Long Short-Term Memory (LSTM) model for forecasting retail demand more efficiently. By incorporating these key macroeconomic indicators, the predictive framework gains superior performance. As hypothesized, the resultant LSTM model, fortified by this infusion of external macroeconomic insights, exhibits superior performance in contrast to a counterpart model constructed without such macroeconomic contextualization.

The literature reviewed in this paper showcases the growing interest in leveraging machine learning models for demand forecasting across diverse industries. As businesses strive for greater forecasting accuracy to inform supply chain decisions, a key opportunity lies in combining the predictive prowess of machine learning with the enriched insights of macroeconomic and consumer data. While Haque (2023) provides evidence of the superior performance of forecasting accuracy for the inclusion of external economic conditions, his finding is based on the results from an LSTM model only. He argues that LSTM is predominantly a superior forecasting model compared to other counterpart models. Therefore, the current literature lacks evidence that this finding from Haque (2023) is broadly generalizable, and valid for other relevant forecasting models. Our study bridges this gap by utilizing various relevant models for forecasting retail demand and comparing performances obtained by including external economic conditions to those that ignore macroeconomic conditions. Findings from our study strongly support the finding from Haque (2023) and provide strong evidence for superior performance of forecasting accuracy for the inclusion of external economic conditions.

## 3. Methodology

**3.1 Data Pre-Processing**

In the phase of data pre-processing, our study embarks on a systematic exploration of historical data provided by Walmart, USA. This dataset spans a temporal expanse of five years and encompasses 3,049 distinct products retailed across ten different stores situated within three states: CA, TX and WI. The dataset encompasses various parameters, including product id, product price, product category, department, store information, promotions, day of the week, and any events coinciding with the product's sale. Importantly, each store within these states maintains an identical inventory of the aforementioned 3,049 products, underscoring the uniformity of the dataset. This extensive dataset comprises a multitude of parameters, including product identifiers, product prices, product categories, department classifications, store particulars, promotional undertakings, day of the week, and any concurrent events linked to the sale of specific products. The calendar dataset provides crucial information pertaining to special events, holidays, promotions, and event types for each day across each state. The product price dataset comprises the pricing information of each product on a daily basis at each store.

Moreover, this study advances towards a more comprehensive analytic phase by incorporating macroeconomic variables of significance, including the Consumer Price Index (CPI), Index of Consumer Sentiment (ICS), and unemployment rates. These economic variables are procured from reputable sources, such as the World Bank's World Development Indicators (WDI) database and the University of Michigan's repository. By utilizing this extensive dataset and incorporating macroeconomic variables, the study seeks to gain comprehensive insights into the relationships between product sales, pricing, day of sale, and the broader economic conditions.

The analysis utilizes a subset of the full dataset filtered for a specific store "CA_1" over the validation period. This extracted dataset provides a representative sample, allowing evaluation of model generalizability. Additionally, it encompasses supplemental information integrated from secondary data sources via merging operations. In particular, attributes capturing calendar effects, macroeconomic indicators, and item pricing data are appended to each record. The inclusion of these external variables, merged from domain-relevant datasets, enriches the feature space to potentially improve model performance. However, simply merging all available data can introduce noise or outdated historical signals. As such, additional filtering focuses the analysis solely on more recent sales data. Transactions prior to a cut-off date are excluded to prevent less relevant signals from interfering with model fitting. This ensures only contemporary sales trends and economic conditions are incorporated during training and validation. Overall, thoughtful data extraction, merging, and filtering techniques procure an informative dataset containing the variables of interest for predicting sales. Trimming outdated observations enables learning only from recent, representative patterns, supporting generalizable insights. The processed dataset provides an appropriate foundation for fitting and evaluating machine learning models.

### 3.2 Feature Engineering

The feature engineering process applies three functions to enrich the dataset with additional attributes intended to improve model performance. The first function introduces temporal features related to the date information. Dates can provide useful insights into seasonality, trends, and external events that may influence the target variable. The second function handles missing values in the price variable to ensure data completeness for analysis. Imputing missing values helps maintain statistical power and prevent biases. The third function calculates lagged and rolling statistical aggregates based on the historical demand data. Incorporating demand lags and averages enables capturing important autoregressive relationships and cyclical effects.

Table 1: List of Independent Variables

| Independent Variables | (1) Descriptions |
| --- | --- |
| demand | No of items sold |
| sell_price | Price of product |
| ics_all | Index of Consumer Sentiment for all goods and services |
| cpi | Consumer Price Index |
| unemp | Unemployment rates |
| lag_t28 | Lagged value of 28 days of items sold |
| rolling_mean_t7 | 7 days rolling average of 28 days lagged values of daily sale |
| rolling_mean_t30 | 30 days rolling average of 28 days lagged values of daily sale |
| rolling_mean_t60 | 60 days rolling average of 28 days lagged values of daily sale |
| rolling_mean_t90 | 90 days rolling average of 28 days lagged values of daily sale |
| rolling_mean_t180 | 180 days rolling average of 28 days lagged values of daily sale |
| rolling_std_t7 | 7 days rolling standard deviation of 28 days lagged values of daily sale |
| rolling_std_t30 | 30 days rolling standard deviation of 28 days lagged values of daily sale |
| Indicator variables | Indicators for months of a year, days of a week, and product ids |

### 3.3 Machine Learning Models

Machine Learning models explored in this study consists of Lasso, Ridge regression, Light Gradient Boosting Machine (LGBM), Extreme Gradient Boosting Machine (XGBM), Decision Tree. We apply aforementioned machine learning techniques on the historical sales data enriched with macroeconomic variables to verify the impact on the macroeconomic variables on the performance of the models. Specifically, we fit those models with only historical sale data and also with historical sale data enriched with macroeconomic variables. We compare the performance of the models developed with and without macroeconomic variables. Model performances are measured in terms of Root Mean Squared Error (RMSE) as well as in terms of Mean Absolute Error (MAE).

## 4. Results and Discussion

Five machine learning models were evaluated on the retail sales dataset - Lasso regression, Ridge regression, XGB, LightGBM, and Decision Tree regression. The models were trained and tested both with and without additional macroeconomic features to assess the impact on predictive accuracy. In Figure 1, we observe a visual comparison between the actual sales and the sales predicted by various machine learning models. Each model's prediction is depicted as a line that fluctuates in relation to the actual sales line. The graph illustrates the impact of incorporating macroeconomic variables on predictive accuracy, as discussed in the study.

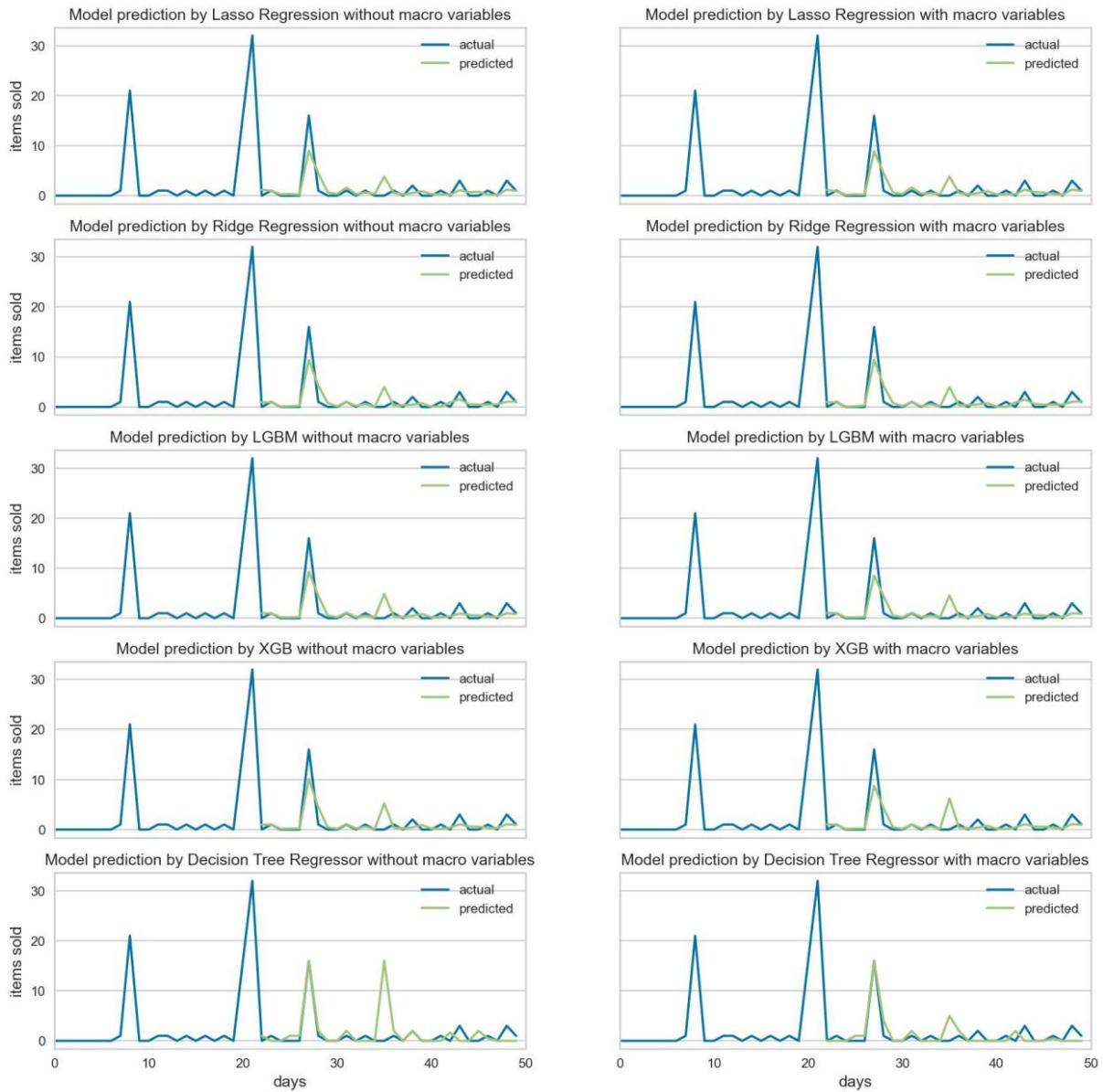

Figure 1: Forecasted values for a single item (HOBBIES_2_084) generated by various models

Models' performances are evaluate using two different measures: Mean Absolute Error (MAE) and Root Mean Squared Error (RMSE) have been calculate using held-out test data. The Lasso model obtains a MAE of 0.886 and RMSE of 1.802 without macro variables. Incorporating macroeconomic features slightly improved performance to MAE of 0.884 and RMSE value of 1.798. For the Ridge regression model, macro variables similarly provids an additional benefit reducing RMSE slightly from 1.739 to 1.738, however, MAEs are approximately equal around 0.845. In contrast, the LightGBM model shows more significant improvements with the addition of macro data. For LightGBM, MAE decreases from 0.849 to 0.847 while RMSE declines from 1.717 to 1.715 when macro variables are added. For XGBM, MAE drops from 0.81 to 0.839 and RMSE drops from 1.721 to 1.715. On the other hand, the Decision Tree model exhibits a slight rise in errors, with MAE risingg from 1.000 to 1.001 and RMSE rising from 2.357 to 2.364 with macro features. Except a few cases, these results in general demonstrate that incorporating external macroeconomic data as model inputs provided mostly modest gains in predictive accuracy across the machine learning models. The tree-based LightGBM model derived the most value from the expanded feature set. The magnitudes of improvement reinforce the importance of relevant domain knowledge in selecting informative input variables. Overall, the findings suggest that macroeconomic indicators meaningfully enhance retail sales forecasting performance for machine learning algorithms.

Table 2: Comparison of Model Performances

| Model | Performance of Models | | | |
|---|---|---|---|---|
| | RMSE | | MAE | |
| | Without Macro Variables | With Macro Variables | Without Macro Variables | With Macro Variables |
| Lasso | 1.80239 | 1.79865 | 0.88665 | 0.88479 |
| Ridge | 1.73869 | 1.73848 | 0.84552 | 0.84567 |
| LGBM | 1.71740 | 1.71504 | 0.84859 | 0.84742 |
| XGBM | 1.72164 | 1.71581 | 0.84126 | 0.83918 |
| Decision Tree | 2.35698 | 2.36479 | 1.00046 | 1.00170 |

Finally, we present an analysis of feature importance utilizing various models to support our claim that macroeconomic variables have explanatory power in explaining future retail demand. As part of this analysis, we identify and present important features that each model considers to be the most important for predicting retail sales. This analysis is based on 100 items instead of all items in a store and results in a total of 131 features. Table 3 highlights the 15 most important features identified by each model out of total 131 features. The set of 15 most important features identified by each model includes one or more macroeconomic variables used in this study. For instance, the Lasso regression model considers the Index of Consumer Sentiment (ICS) as one of the most important features. Similarly, Ridge regression identifies CPI as one of the important features out of 131 features. Similarly, each list of important features identified by other models also includes one or more external economic condition-related variables. Overall, the demand forecasting model considers external economic conditions as important determinants for future demand.

Table 3: Important Features Identified by Models

| Order of Importance | Lasso | Ridge | LGBM | XGB | Decision Tree |
|---|---|---|---|---|---|
| 1 | rolling_mean_t30 | item_id_HOBBIES_1_048 | rolling_mean_t30 | rolling_mean_t7 | rolling_mean_t30 |
| 2 | rolling_mean_t7 | item_id_HOBBIES_1_008 | rolling_mean_t7 | lag_t28 | rolling_mean_t60 |
| 3 | rolling_mean_t60 | item_id_HOBBIES_1_019 | rolling_std_t7 | rolling_std_t30 | rolling_std_t30 |
| 4 | rolling_mean_t180 | item_id_HOBBIES_1_016 | rolling_mean_t90 | rolling_mean_t7 | rolling_mean_t90 |
| 5 | lag_t28 | item_id_HOBBIES_1_030 | rolling_mean_t180 | rolling_std_t7 | rolling_std_t7 |
| 6 | sell_price | item_id_HOBBIES_1_015 | Ics_all | rolling_mean_t60 | sell_price |

| 7 | ICS_ALL | item_id_HOBBIES_1_032 | Sell_price | rolling_mean_t90 | rolling_mean_t7 |
| --- | --- | --- | --- | --- | --- |
| 8 | -- | item_id_HOBBIES_1_043 | cpi | rolling_mean_t180 | rolling_mean_t180 |
| 9 | -- | item_id_HOBBIES_1_004 | Week_day_5 | ICS_ALL | lag_t28 |
| 10 | -- | week_day_5 | Week_day_6 | sell_price | item_id_HOBBIES_1_048 |
| 11 | | rolling_mean_t30 | item_id_HOBBIES_1_015 | cpi | ICS_ALL |
| 12 | | item_id_HOBBIES_1_014 | item_id_HOBBIES_1_048 | Week_day_6 | cpi |
| 13 | | week_day_6 | item_id_HOBBIES_1_016 | Week_day_5 | week_day_3 |
| 14 | | cpi | Week_day_3 | Week_day_0 | week_day_5 |
| 15 | | month_1 | item_id_HOBBIES_1_016 | Week_day_1 | week_day_6 |

# 5. Conclusion

Our study stands at the juncture of two critical domains: demand forecasting within the retail industry and the profound impact of macroeconomic variables on consumer behavior. Through a meticulously designed approach, we have seamlessly integrated macroeconomic indicators—namely CPI, ICS, and unemployment rates—into historical sales data. Our findings fortify the findings from a previous study in the current literature in a more generalizable manner – that by including external macroeconomic data into our models we can improve the predictive accuracy, and that these findings obtained from an LSTM model in the current literature are also valid for other relevant models in the machine learning horizon. The significance of our findings resonates deeply within the retail sector, where data-driven decisions underpin success. Armed with insights generated from this study, businesses can now more astutely fine-tune their inventory management strategies and refine their supply chain decisions. Our study also highlights that we can gain modest to substantial improvements across various machine learning models by including external economic variables and among these models, the tree based LightGBM stands out as the one benefiting the most from this approach. These findings, grounded in empirical analysis, hold the potential to revolutionize the operational landscape, driving efficiency, reducing costs, and bolstering competitive edge. As the retail realm continues its transformative journey, the confluence of analytical machine-learning capabilities and the nuanced context provided by macroeconomic indicators offers a promising horizon. This amalgamation, demonstrated through our research, equips organizations with a potent toolkit to forge ahead with superior predictive accuracy. This clarity has the potential to reshape the future of how retail works.

# Declarations

**Funding**
This research is a result of personal efforts of authors, and no institutional funding is applicable for this study.
**Conflicts of interest/Competing interests**
Authors have no conflict of interest.
**Data availability**